\documentclass[a4paper,twoside]{article}

\usepackage{epsfig}
\usepackage{subcaption}
\usepackage{calc}
\usepackage{amssymb}
\usepackage{amstext}
\usepackage{amsmath}
\usepackage{amsthm}
\usepackage{multicol}
\usepackage{pslatex}
\usepackage{apalike}
\usepackage[bottom]{footmisc}
\usepackage{SCITEPRESS}     

\begin{document}
\title{DEff-GAN: Diverse Attribute Transfer for Few-Shot Image Synthesis}
\author{\authorname{Rajiv Kumar\sup{1}, and G. Sivakumar\sup{2}}
\affiliation{\sup{1,2}Department of CSE, IIT Bombay, Mumbai, INDIA} \email{\sup{1}rajiv@cse.iitb.ac.in, \sup{2}siva@iitb.ac.in}}

\abstract{Requirements of large amounts of data is a difficulty in training many GANs. Data efficient GANs involve fitting a generator's continuous target distribution with a limited discrete set of data samples, which is a difficult task. Single image methods have focused on modelling the internal distribution of a single image and generating its samples. While single image methods can synthesize image samples with diversity, they do not model multiple images or capture the inherent relationship possible between two images. Given only a handful number of images, we are interested in generating samples and exploiting the commonalities in the input images. In this work, we extend the single-image GAN method to model multiple images for sample synthesis. 
We modify the discriminator with an auxiliary classifier branch, which helps to generate wide variety of samples and  to classify the input labels. Our \textbf{D}ata-\textbf{Eff}icient \textbf{GAN} (\textbf{DEff-GAN}) generates excellent results when similarities and correspondences can be drawn between the input images/classes.}
\onecolumn \maketitle \normalsize \vfill
\section{\uppercase{Introduction}}
\label{sec:introduction}
Most of the modern deep learning based methods depend on large datasets and need long training times \cite{BigBiGAN}, \cite{Karras2019_stylegan2} for achieving high performance and state-of-the-art results. The trend still continues and is observed even in some of the few-shot learning tasks \cite{FUNIT2019}. However, there are use cases and scenarios where obtaining even a handful number of images is difficult due to reasons of privacy, security and ethical reasons. Though Generative Adversarial Networks (GANs) are able to generate realistic images of high quality \cite{BigBiGAN}, \cite{rajiv_VISAPP21}, \cite{Karras2019_stylegan2}, this is possible with the availability of large and diverse training datasets \cite{MIS_CD} that prevents memorization problems. In most cases, the amount of data needed for training or adapting a GAN is in the order of hundreds, if not in thousands, thereby leaving no purpose in generating more of the same data. 
In the few-shot realm, when GANs are trained directly with small datasets, it leads to severe quality degradation or memorization issues or both. Therefore, it becomes essential to prevent mode collapse and overfitting to generate samples with diversity. 
\par
Recently, there has been interest in single image generative models to synthesize image samples of various scales and sizes. Single-image GAN models \cite{InGAN_2019}, \cite{SinGAN_2019}, \cite{ConSinGAN_2020}, \cite{OneShotGAN_2021} have overcome the overfitting and mode collapse issues by learning from the internal distribution of patches from a single image. However, the synthesized image samples make little to no sense when they lack coherence. Efforts to improve the diversity in a few-shot setting leads to artifacts, poor realism, and incoherency in images. 
With only a single image modelled by a GAN, there are applications like image super-resolution, harmonization, etc., which is possible by using the patches from the input image itself. Modelling multiple images can result in generalization as well as learning the underlying semantic relations between the images. This leads to potential for learning the relation between the patches from multiple images that opens up the possibilities of style transfer, content transfer, image compositing, image blending, etc. 
In a few shot scenario, novel sample synthesis is possible by transferring visual attributes like color, tone, texture or style from one image to another and by combining features from different inputs. For unsupervised image synthesis, the visual attributes can come from different images without any guidance on how the features should be combined. 
\par
In this paper, we illustrate that single-image GANs can be adapted for multi-class image synthesis in a few-shot setting for similar classes. Lets consider the case of two face images, where correspondences can be drawn between common facial features like eyes, nose, lips, hair, etc. These correspondences can give rise to similarities and relations at the local patch level, which can be leveraged for novel sample synthesis. 
For this, we propose changes to existing single image GAN \cite{ConSinGAN_2020} to adapt it for multi-class few-shot image synthesis.  Previous methods like SinGAN and ConSinGAN had focused only in the generation of samples of a single image. 
We propose changes to generate samples with attributes from multiple images by the use of an auxiliary classifier branch for the discriminator, to output the class probabilities in addition to the real/generated labels. The discriminator objective then includes the classifier loss that minimizes the cross entropy loss between the labels of generated images and the class labels. We also modify the training procedure for modelling multiple images and to speed up the training, while single-image GAN methods generate a single sample every time. As a result, for images with similar semantics and underlying content, our method synthesizes novel samples in a few shot setting. In the case of face images and textures, our method can result in diverse sample synthesis generating hundreds of variations while retaining the semantics, from a single image of two different faces images. 
The paper contributions are as follows:
\begin{itemize}
\item We introduce \textbf{DEff-GAN}, a pretraining-free few-shot image synthesis method by adapting single-image GAN methods for multiple images for diverse novel sample synthesis.
\end{itemize}
We briefly explain the Related works in Section \ref{Sec:RELATED WORKS}, Methodology in Section \ref{Sec:METHODOLOGY}, Implementation details in Section \ref{Sec:IMPLEMENTATION}, Experiments and evaluation in Section \ref{Sec:EXPERIMENTS}, 
Results and analysis in Section \ref{Sec:RESULTS} and Conclusion and future scope in Section \ref{sec:conclusion}.
\section{RELATED WORKS}
\label{Sec:RELATED WORKS}
There are various approaches for few-shot generation, from direct training of few-shot image datasets to few-shot test time generalization. In the former case, a generative model is trained directly on a small dataset with a handful of images without adapting a pre-trained model or training on large number of base categories. In the latter case, generative models are trained on a set of base categories for long training schedules and later applied to novel categories with optimization \cite{FIGR_2019}, \cite{DAWSON_2020} or finetuning. In some cases, there is no optimization involved when using fusion-based methods \cite{MatchingGAN_2020}, \cite{F2GAN_2020}, \cite{LoFGAN_2021} or transformation-based methods \cite{DeltaGAN_2022}, \cite{Ding2022AttributeGE}. One way for knowledge transfer is to use pre-trained models from related domains and adapting it using only a few input images. However, the resulting network can still be large which can easily overfit to the data since the number of samples is very less. 
\par
Recent works \cite{zeroshotSR_shocher2017} perform various tasks \cite{MorphGAN2020}, \cite{RepurposeGANs_2021} using very few data samples \cite{DLSSR-review} and even from a single image \cite{SinGAN_2019}, \cite{InGAN_2019}, \cite{ConSinGAN_2020}. 
We briefly explain the similarities and differences of single image GAN methods and their drawbacks. InGAN \cite{InGAN_2019} focuses on the completeness and coherence of the generated images with an encoder-encoder architecture that generates sample images of various shapes, sizes and aspect ratios. 
SinGAN \cite{SinGAN_2019} is a single image based GAN framework for image harmonization, image editing, super-resolution tasks, etc. 
ConSinGAN \cite{ConSinGAN_2020} takes one step further by improving the speed of training of SinGAN and also improves on the number of stages required for generating an image of required resolution. InGAN and rcGAN \cite{rcGAN} learns the distribution of image patches of multiple images in the same model and fills in the patches from the training image for image manipulations and downstream tasks. SA-SinGAN \cite{Chen2021_SA_Singan} uses self-attention mechanism in a single image model to improve the image quality by obtaining the global structure and also improves the training time. 
While the above methods generate appealing results, most single image methods have not been adapted or illustrated to work with multiple images/classes. 
\par
In the setting of learning from a single video, One-shot GAN~\cite{OneShotGAN_2021} uses a two-branch discriminator to assess the internal content from the scene layout with separate content and layout branches.
In a few-shot setting, one method~\cite{FastGAN_2021} works with dataset sizes up to 100 images but fails for fewer images ($\leq$ 10) in terms of sample diversity, as generated samples become limited to input image reconstructions. Another method \cite{FSGANadapt} can adapt a pre-trained GAN with as few as 10 images by learning cross-domain correspondences. However, it is difficult to find a GAN pre-trained on related domains and only the style parameters of the pre-trained GAN are altered, which prevents capturing of the underlying semantics of the target domain. 
For pretraining-free few-shot image synthesis, one method \cite{mixdl_2021} proposes a mixup-based distance regularization on the feature space of both the generator and discriminator to enhance both fidelity and diversity. 
\section{METHODOLOGY}
\label{Sec:METHODOLOGY}
\subsection{Problem Formulation}
For the few-shot image synthesis task, we consider two images, $x_1$ and $x_2$ belonging to the same class as the base case. The goal is to learn a generative model that can generate samples of large diversity with visual attributes from the two input images. Similarly, for the multi-class image synthesis problem, we consider a set of $k$ images, $\{x_1$, $x_2$... $x_k\}$ belonging to the related classes. Given a set of $k$ images, which is usually a small number (k $<$ 5), our goal is to learn a model that can generate samples of the $k$ related classes using a single image of each class, for the image synthesis problem. 
\subsection{Proposed Framework}
For modelling a few number of images using a generative model, training a lightweight model is preferable than adapting a pretrained model that was trained on a large dataset. 
Single-image based sample synthesis is generally based on progressive growing based architectures with multi-stage and multi-resolution training. This gives greater control over the image generation process and its quality in comparison to end-to-end training of the whole network, which otherwise may also overfit to the input images. The receptive fields at varying scales are captured by a cascade of patch-GANs with progressive field of view to capture the patch distributions at that scale and by scaling up through image sizes.
An unconditional generative model is learned as a growing generator by adding new layers, keeping the previous stages frozen or trained at small learning rates. To this end, we detail the design and details of our framework for one-shot multi-class image synthesis and few-shot image synthesis.
\subsection{Design}
\begin{figure}
\centering
\includegraphics[scale=0.085]{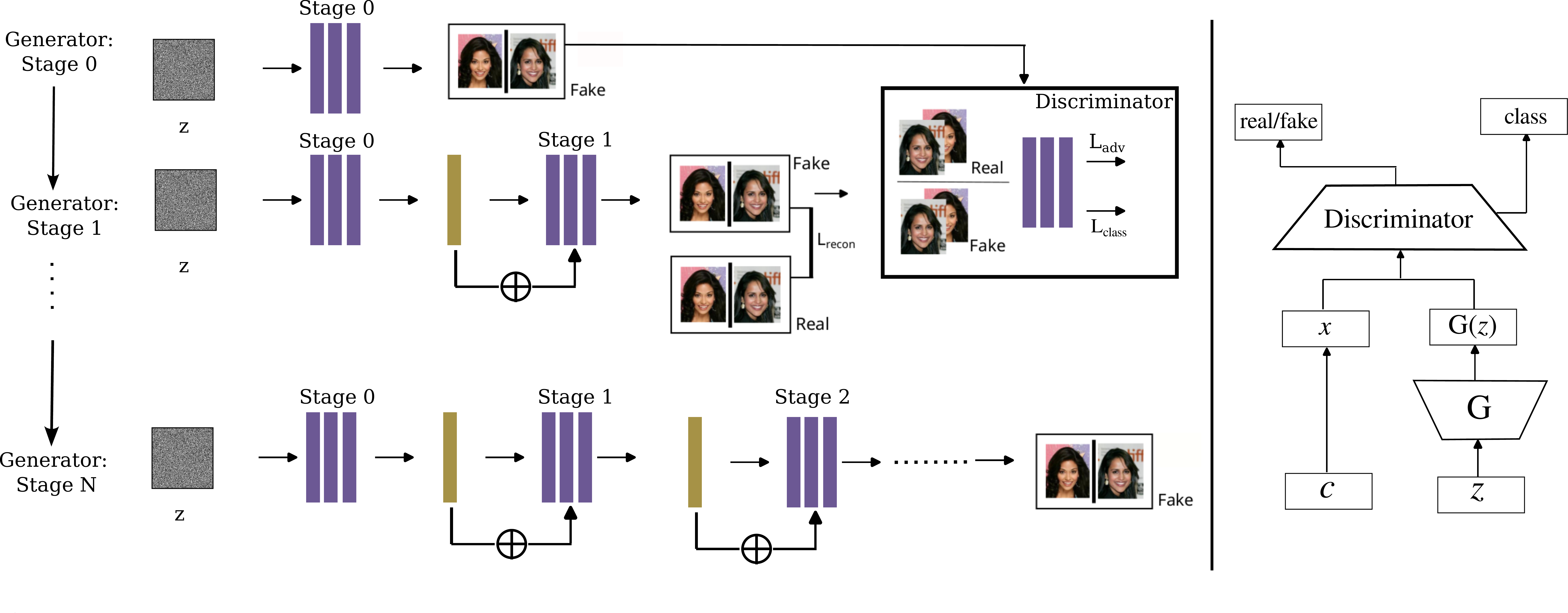}
\caption{Layout diagram illustrating the different stages and the relation between the real and generated images. The right section illustrates the relation between the various images, the generator and discriminator networks.}
\label{fig:layout_diagram}
\end{figure}

In principle, we could adapt the architecture of SinGAN \cite{SinGAN_2019} or that of ConSinGAN \cite{ConSinGAN_2020}. We adapt ConSinGAN architecture for our method due to faster training speeds and concurrent training of multiple stages. Hence, our method has commonalities in terms of design, architecture and implementation with ConSinGAN \cite{ConSinGAN_2020}. Also, we use features in our method between the generator stages, rather than image outputs from the previous stage generators. 
We employ a pyramid of fully convolutional patch-GANs, which consists of generators stages $\{G_{N}, G_{N-1}... G_0\}$ and discriminators $\{D_{N}, D_{N-1}... D_0\}$. We associate each generator stage $G_i$ from $\{G_{N}, G_{N-1}... G_0\}$ with a discriminator $D_i$ from $\{D_{N}, D_{N-1}... D_0\}$, for i $\epsilon$ $\{N, N-1... 0\}$ (refer Figure \ref{fig:layout_diagram}). Generator stage $G_0$ corresponds to the image of coarsest scale, while the generator stage $G_N$ corresponds to generator dealing with the finest details. 
\par
Let's consider the training of generator at stage $i$, for $i < N$. During the training stage $i$, the generator stage $G_i$ and discriminator $D_i$ are trained. The generator training at any stage $i$ requires only fixed noise maps and noise samples to the unconditional generator at the coarsest scale, with features from the lower stages propagated to the higher stages. Once the generator at a scale $i$ is trained completely, then training proceeds to the generator stage ${i+1}$ and so on. There are different set of images involved in training at any scale $i$, i.e. real images and generated images at scale $i$. The growing generator learns by adversarial training by generating images and by minimizing the reconstruction loss of generated images to real images. 
\subsection{Objective Function}
For adversarial training, we consider a set of real and fake images, which correspond to the dataset images and generated images correspondingly. In our method, the discriminator is modified to have an auxiliary classifier branch to classify the inputs in addition to the discriminator's real/fake label that helps in adversarial learning. However, our generator is different from that of AC-GAN, since it is dependent only on the noise samples and independent of the class labels, while the generator used in AC-GAN takes the class label along with the noise samples while generating samples. We do not condition the generator on an input image or class label and hence our generator is unconditional, while our discriminator has an auxiliary classifier branch. The growing generator G is a lightweight iterative optimization based network that learns to map randomly sampled noise $z$ belonging to $Z$ to the output space of images, ${G: Z->X}$. $l(.)$ is a distance metric in the image space which can either belong to ${l_1}$ or ${l_2}$ norm. 
We consider Mean Squared Error (MSE) pixel reconstruction loss enforced between the real images and reconstructed images for samples generated using fixed noise maps, as given in Equation \ref{Eq:L2 reconstruction}. The generator's objective is to fool the discriminator into identifying the generated images as real and to reduce the reconstruction loss. The generator objective involves an adversarial loss and reconstruction loss, as given in Equation \ref{Eq:Adversarial_n_recon}. More importantly, we do not have an adversarial or supportive classifier loss enforced as a part of generator's objective.
\begin{equation}
\mathcal{L}_{\textit{rec}}(G_n) = \vert\vert G_n(z) - x_n \vert\vert^2_2.
\label{Eq:L2 reconstruction}
\end{equation}
\begin{equation}
\underset{G_n}{min}\ \underset{D_n}{max}\ \mathcal{L}_{\textit{adv}}(G_n, D_n) + \alpha \mathcal{L}_{\textit{rec}}(G_n).
\label{Eq:Adversarial_n_recon}
\end{equation}
\noindent
The discriminator objective function consists of two parts: the log-likelihood of the correct source, $L_S$ given in Equation \ref{discriminator_source_loss} and the log-likelihood of the correct class, $L_C$,  given in  Equation \ref{discriminator_class_loss}. The discriminator gives a probability distribution over sources (real/generated), $P(S|X)$ and a probability distribution over the class labels, $P(C|X)$ = $D(X)$. 
\begin{multline} \label{discriminator_source_loss}
L_S = E[\log P(S=real~|~X_{real} )] + \\ E[\log P(S=fake~|~X_{fake})].
\end{multline}
\begin{multline}
\label{discriminator_class_loss}
L_C = E[\log P(C = c ~~|~ X_{real} )] + \\ E[\log P(C = c ~|~ X_{fake})].
\end{multline} 
\par
The discriminator is trained to maximize $L_S$ and $L_C$, while $G$ is trained to maximize $L_S$. For the real input images, cross entropy loss is enforced between class labels of the randomly ordered training batch and the classifier outputs of the discriminator. For the fake images, it is desired to have attributes from multiple inputs for attribute transfer and hence we assign class labels in a random fashion resulting in generated images to take attributes from other classes. The discriminator is provided with input images labelled as real and generated images labelled as fake.
Gradient penalty is computed between the real and the fake images and the gradients are back propagated using WGAN-GP \cite{WGAN_GP-GulrajaniAADC17} adversarial loss. To prevent mode collapse and to capture the complete set of real images, we consider each training batch to comprise of the whole set of input images. Considering batch sizes smaller than the full set of real images may lead to non-capturing of all modes. Also, the input images are fixed before the critic operations and shuffled in each critic iteration in a random order. To be memory-efficient while handling multiple images, the fixed maps are generated only for the coarsest scale, while previous methods have considered a pyramid of fixed maps for each image corresponding to each scale. Consequently, we abstain from adding noise after each stage with upsampling step after observing that it has little to no effect during our method implementation. 
\section{IMPLEMENTATION}
\label{Sec:IMPLEMENTATION}
The implementation of our method involves a growing generator and a pyramid of discriminators. The generator and discriminator start with the same number of convolutional layers. As training proceeds, the generator is progressively grown by concatenating the latest stage  that captures the patch distribution at that scale to the previously trained stages. 
We suggest concurrent training of at least two stages and the learning rates are exponentially decayed along the stages so to fine-tune the network weights of previous stages. 
We use a pyramid of the scaled real image for each training stage for each image. We randomly select one of the $k$ images in each iteration and the associated pyramids with it, while training on that image. A fixed noise map is a random noise map that is assigned at the beginning of training and fixed for each train image for the coarsest scale
and used for reconstruction of input images. For WGAN-GP, the number of critic iterations per generator iteration is usually fixed between 3 and 5. 
Differentiable Augmentation \cite{DiffAugment_2020} is an augmentation technique that improves the data efficiency of GANs for both unconditional and class-conditional generation, by imposing various types of differentiable augmentations on both real and fake samples. We observe that differentiable augmentation with color helps to improve the quality of the generated images, while cutout and translation have detrimental effects in some cases.

\begin{figure}
\centering
\includegraphics[scale=0.11]{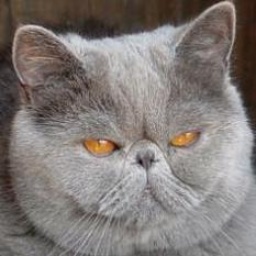}
\includegraphics[scale=0.11]{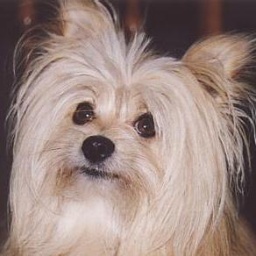}
\includegraphics[scale=0.3]{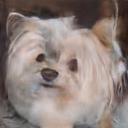}
\includegraphics[scale=0.3]{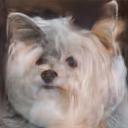}
\includegraphics[scale=0.3]{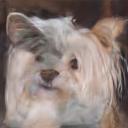}
\includegraphics[scale=0.3]{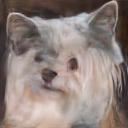}
\includegraphics[scale=0.3]{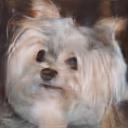}

\includegraphics[scale=0.11]{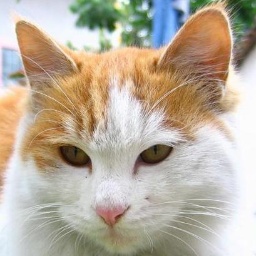}
\includegraphics[scale=0.11]{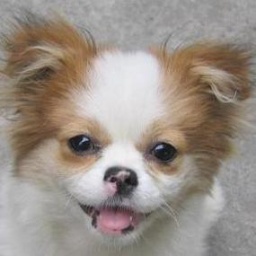}
\includegraphics[scale=0.3]{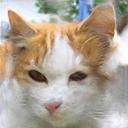}
\includegraphics[scale=0.3]{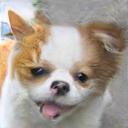}
\includegraphics[scale=0.3]{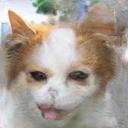}
\includegraphics[scale=0.3]{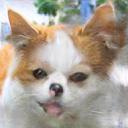}
\includegraphics[scale=0.3]{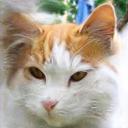}

\includegraphics[scale=0.11]{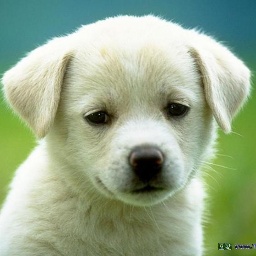}
\includegraphics[scale=0.11]{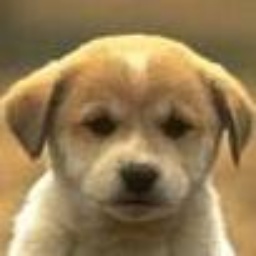}
\includegraphics[scale=0.3]{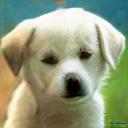}
\includegraphics[scale=0.3]{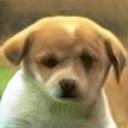}
\includegraphics[scale=0.3]{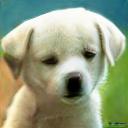}
\includegraphics[scale=0.3]{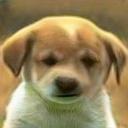}
\includegraphics[scale=0.3]{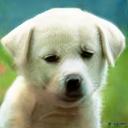}
\caption{Multi-class image synthesis on cats and dog classes. The leftmost two columns are the real images while the rest of the images are generated images.}
\label{fig:Dog2cat_p3_v11}
\end{figure}
\begin{figure}
\centering
\includegraphics[scale=0.182]{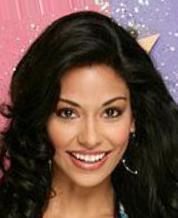}
\includegraphics[scale=0.25]{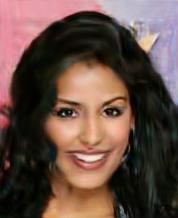}
\includegraphics[scale=0.25]{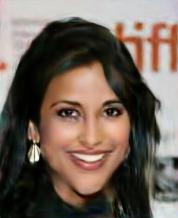}
\includegraphics[scale=0.25]{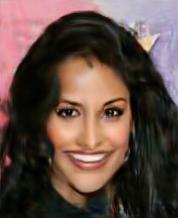}
\includegraphics[scale=0.25]{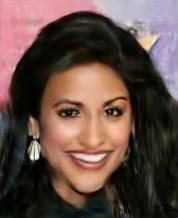}
\includegraphics[scale=0.25]{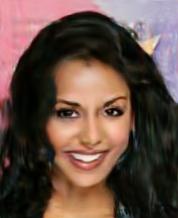}
\includegraphics[scale=0.182]{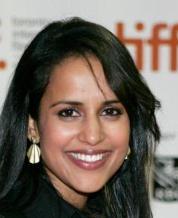}
\includegraphics[scale=0.25]{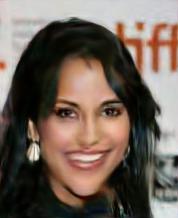}
\includegraphics[scale=0.25]{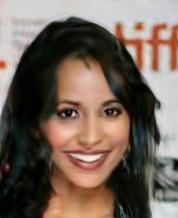}
\includegraphics[scale=0.25]{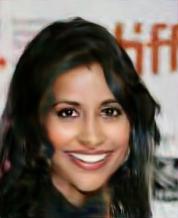}
\includegraphics[scale=0.25]{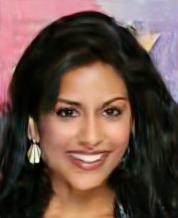}
\includegraphics[scale=0.25]{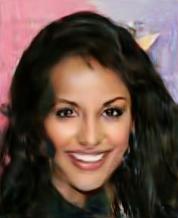}
\caption{Few shot image synthesis of two face images. The leftmost column has the inputs and the rest of the images are generated images (256 x 256).}
\label{fig:one_shot_face_composition_p9}
\end{figure}

\subsection{Architecture}
We train our method with the following hyper-parameters. When multiple generator stages are concurrently trained, we use a learning rate scaling of 0.5 between any stage and its previous stages. The number of training stages can be varied between 6 to 8 for training images up to dimensions 256 x 256. 
The number of input channels are 3 and the number of filters in the convolution layers can 
be 64 or 128 filters for most cases. Using larger number of filters come at the expense of more GPU memory consumption. We use prelu as the activation function and $\alpha$, the weight for reconstruction loss as 10. We use Adam optimizer with betas of 0.5 and 0.999. While training, we explicitly set the learning rate of discriminator at 0.00025, half as that of generator at 0.0005. 
We use multi-step learning rate scheduler with a gamma value of 0.1 and milestones as 0.8 times the number of images times the number of iteration per image. The penultimate and the last stage can be trained for extended iterations to further improve the quality of generated images. The number of convolutional layers in each stage can be varied from 3 to 6 depending on the number of images that are modelled. 
\begin{figure}
\centering
\includegraphics[scale=0.4]{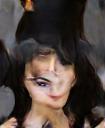}
\includegraphics[scale=0.4]{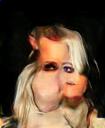}
\includegraphics[scale=0.4]{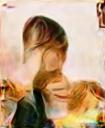}
\caption{Generated samples (128x128) from ConSinGAN on modelling two inputs for different sets of faces. Samples are affected by mode collapse and are incoherent.}
\label{fig:mode_collapse}
\end{figure}
\section{EXPERIMENTS AND EVALUATION}
\label{Sec:EXPERIMENTS}
Most of the related few-shot generative methods require pre-training on a large number of base classes and long training schedules, since they focus on few-shot test-time generalization. We have not considered these methods as baselines, since they benefit from prior-knowledge from previously seen data and have an unfair advantage in a true few-shot setting. We choose mixup-based distance learning method \cite{mixdl_2021} as the baseline for comparing our method for few-shot image generation. 
\subsection{Datasets}
We consider images from multiple datasets to illustrate the flexibility of our method. The inputs/classes used in our experiments belong to human faces, cat and dog faces, etc. The face images are sourced from celebA~\cite{celeba} and anime datasets. We source the flower images from the Oxford 102-flowers dataset for the few-shot image synthesis task. We also considered a subset of 10 images from 100-shot-Obama dataset for the few-shot image synthesis.
\subsection{Evaluation Metric}
Image synthesis quality of the GAN generated samples are usually evaluated by Inception Score (IS), Frechet Inception Distance (FID) and Learned Perceptual Similarity (LPIPS). We assess the quality of the generated images using LPIPS and SIFID \cite{SinGAN_2019} for evaluating our method. While, SIFID compares one input image against the set of generated images, FID metric is designed to compute the distance between comparable number of real and generated images. However, the number of input images are limited in few-shot multi-class image synthesis while the generated samples are diverse and large in number. A recent work \cite{OneShotGAN_2021} points out that SIFID tends to penalize the diversity and favors overfitting and hence may not be the best metric to evaluate diverse images. The diversity of the generated samples can be measured by LPIPS metric \cite{LPIPS_DosovitskiyB16}. 
\subsection{Experiments}
Unlike single image GANs, the images generated by our method in one-shot multi-class image synthesis/few-shot image synthesis tasks have features and attributes from multiple input classes/images. Since each generated sample has attributes from multiple inputs/classes, we compute SIFID metric on the complete set of generated images against each input image.  
For all FID and SIFID computation, 100 images were generated against two, three, five or ten input images for our method. For our baseline (Mixdl), the checkpoints at 10K intervals were used to generate 5000 images from which 1000 images were considered for computing the above metrics. For different experiments, the images of size 128 or 256 were generated, while all experiments that compare our method with Mixdl \cite{mixdl_2021}  are compared on image size of 256 x 256.
\par
\begin{figure}
\centering
\includegraphics[scale=0.065]{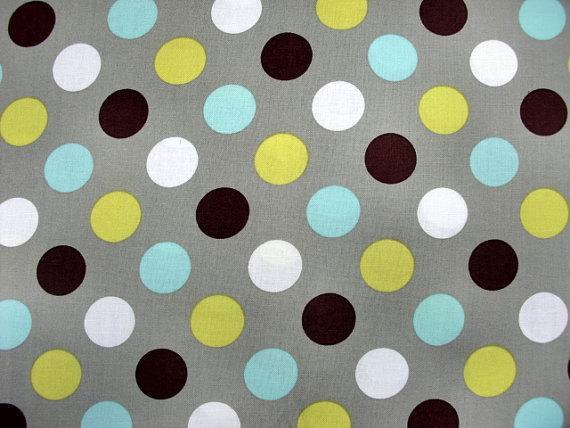}
\includegraphics[scale=0.0756]{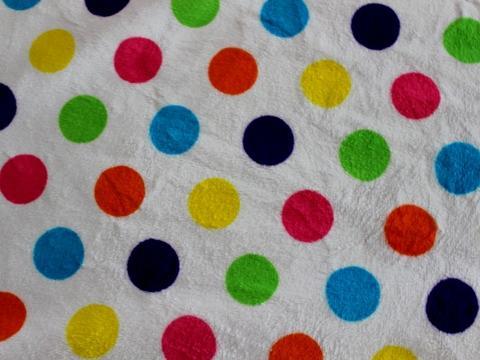}
\includegraphics[scale=0.2]{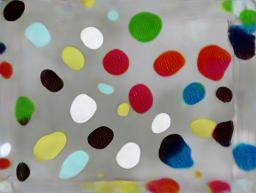}
\includegraphics[scale=0.2]{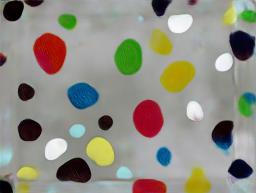}
\includegraphics[scale=0.2]{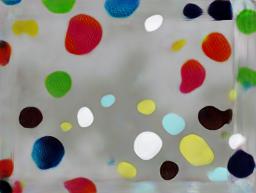}
\includegraphics[scale=0.2]{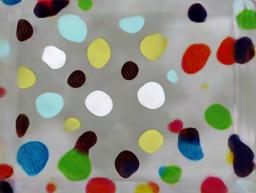}
\includegraphics[scale=0.2]{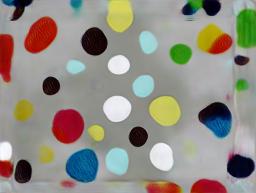}
\includegraphics[scale=0.2]{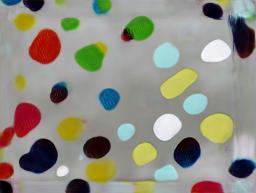}
\includegraphics[scale=0.2]{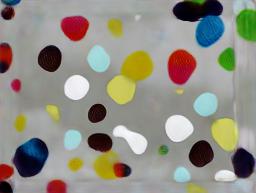}
\includegraphics[scale=0.2]{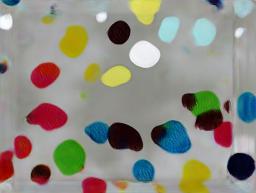}
\includegraphics[scale=0.2]{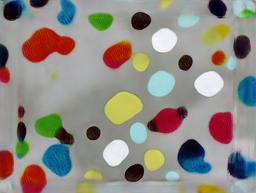}
\includegraphics[scale=0.2]{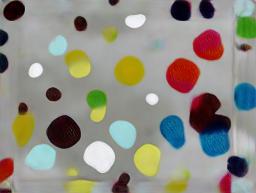}
\includegraphics[scale=0.2]{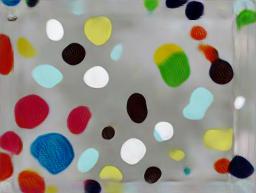}
\includegraphics[scale=0.2]{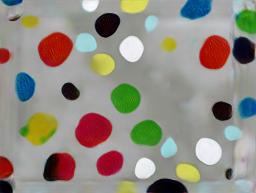}
\includegraphics[scale=0.2]{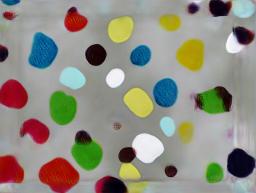}
\caption{Few-shot image synthesis on polka dot texture class. The top row leftmost two images are the inputs and the rest are generated images (256x256).}
\label{fig:few_shot_synthesis_polka_dotted_256}
\end{figure}
\begin{figure}
\centering
\includegraphics[scale=0.2]{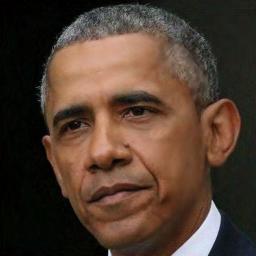}
\includegraphics[scale=0.2]{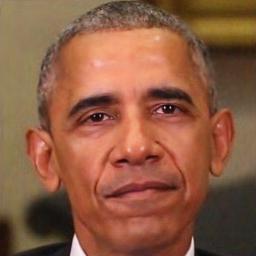}
\includegraphics[scale=0.2]{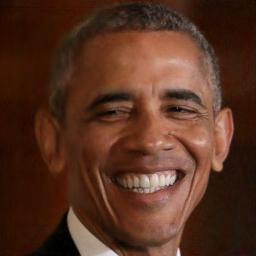}
\includegraphics[scale=0.2]{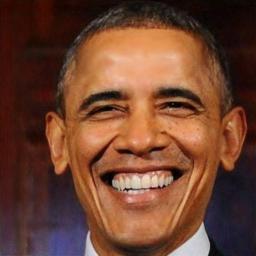}
\includegraphics[scale=0.2]{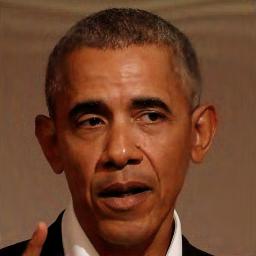}
\includegraphics[scale=0.2]{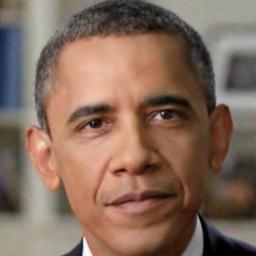}
\includegraphics[scale=0.2]{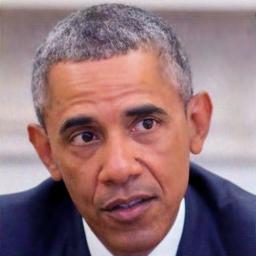}
\includegraphics[scale=0.2]{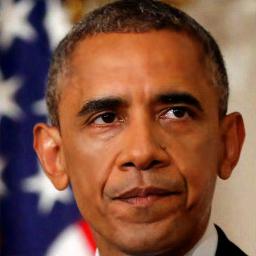}
\includegraphics[scale=0.2]{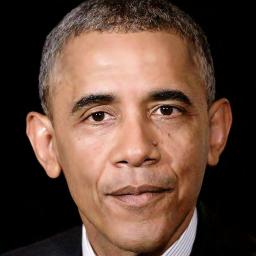}
\includegraphics[scale=0.2]{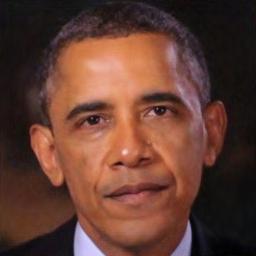}
\includegraphics[scale=0.4]{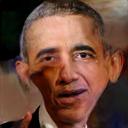}
\includegraphics[scale=0.4]{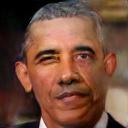}
\includegraphics[scale=0.4]{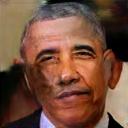}
\includegraphics[scale=0.4]{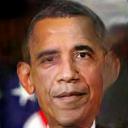}
\includegraphics[scale=0.4]{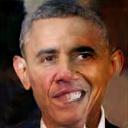}
\includegraphics[scale=0.4]{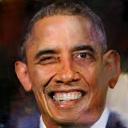}
\includegraphics[scale=0.4]{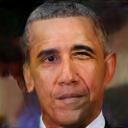}
\includegraphics[scale=0.4]{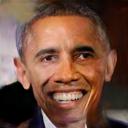}
\includegraphics[scale=0.4]{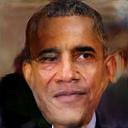}
\includegraphics[scale=0.4]{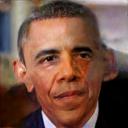}
\includegraphics[scale=0.4]{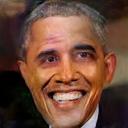}
\includegraphics[scale=0.4]{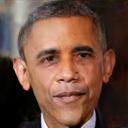}
\includegraphics[scale=0.4]{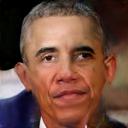}
\includegraphics[scale=0.4]{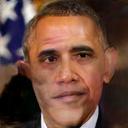}
\includegraphics[scale=0.4]{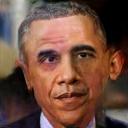}
\includegraphics[scale=0.4]{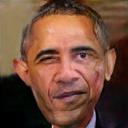}
\includegraphics[scale=0.4]{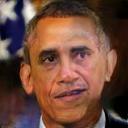}
\includegraphics[scale=0.4]{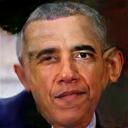}
\includegraphics[scale=0.4]{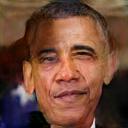}
\includegraphics[scale=0.4]{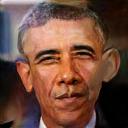}
\caption{Few-shot synthesis on ten selected Obama face images. The top two rows images are the input images, while the rest of the images are generated images (128x128).}
\label{fig:obama-10}
\end{figure}
\begin{figure}
\centering
\includegraphics[scale=0.2]{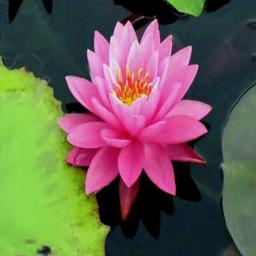}
\includegraphics[scale=0.2]{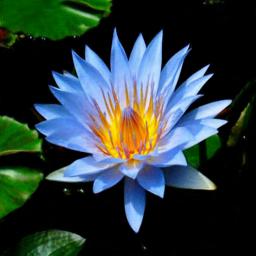}
\includegraphics[scale=0.2]{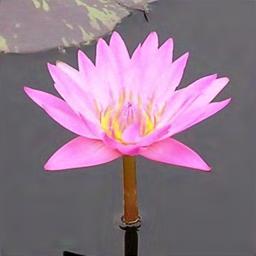}
\includegraphics[scale=0.2]{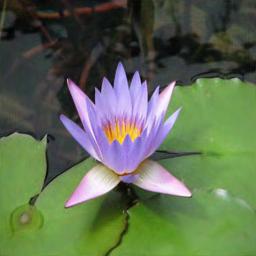}
\includegraphics[scale=0.2]{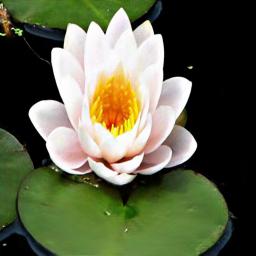}
\includegraphics[scale=0.2]{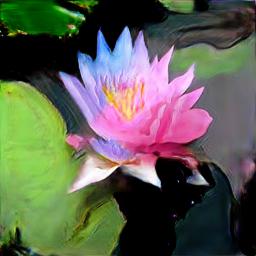}
\includegraphics[scale=0.2]{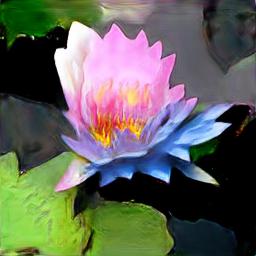}
\includegraphics[scale=0.2]{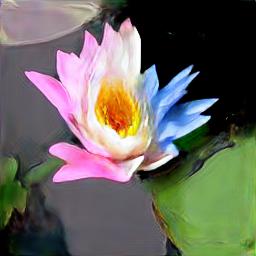}
\includegraphics[scale=0.2]{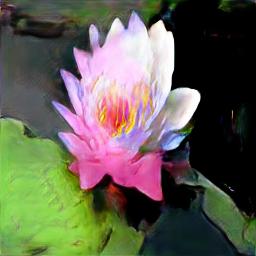}
\includegraphics[scale=0.2]{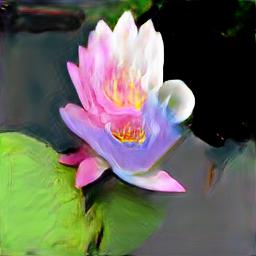}
\includegraphics[scale=0.2]{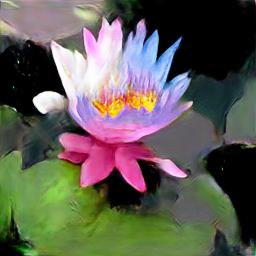}
\includegraphics[scale=0.2]{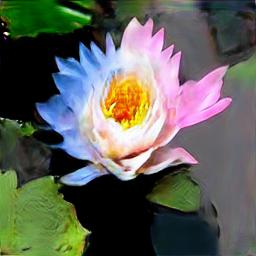}
\includegraphics[scale=0.2]{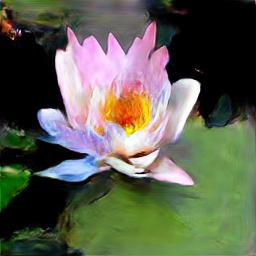}
\includegraphics[scale=0.2]{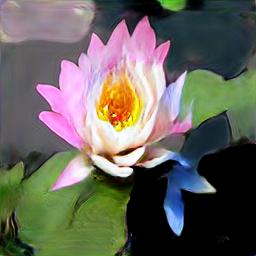}
\includegraphics[scale=0.2]{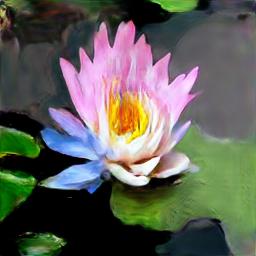}
\includegraphics[scale=0.2]{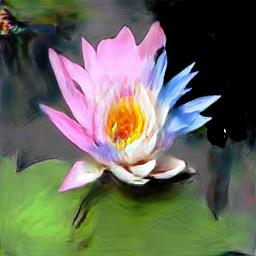}
\includegraphics[scale=0.2]{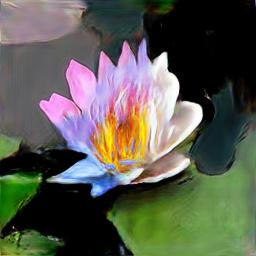}
\includegraphics[scale=0.2]{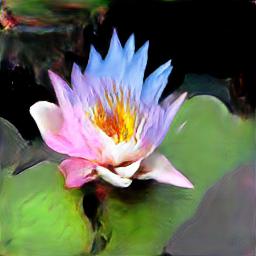}
\includegraphics[scale=0.2]{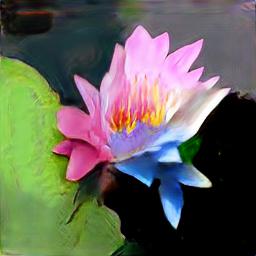}
\includegraphics[scale=0.2]{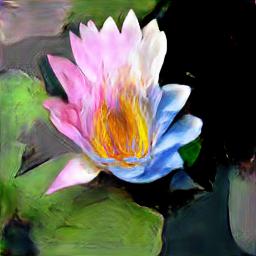}
\includegraphics[scale=0.2]{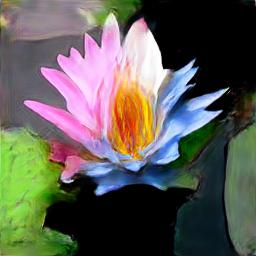}
\includegraphics[scale=0.2]{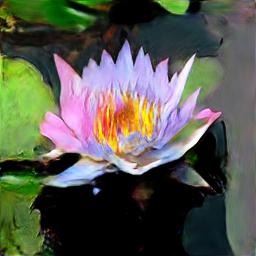}
\includegraphics[scale=0.2]{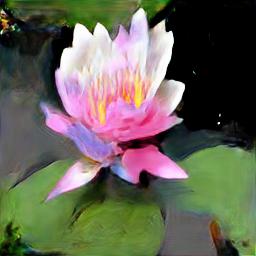}
\includegraphics[scale=0.2]{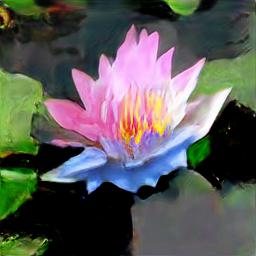}
\includegraphics[scale=0.2]{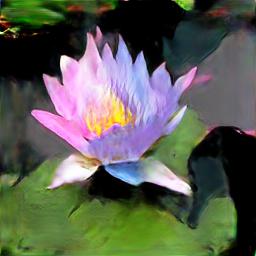}
\caption{Few-shot image synthesis on five flower images. The first row images are the input images, while the rest of the images are generated images (256x256).}
\label{fig:flower-101}
\end{figure}
\begin{figure}
\centering
\includegraphics[scale=0.125]{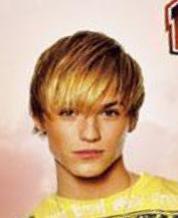}
\includegraphics[scale=0.125]{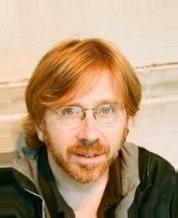}
\includegraphics[scale=0.30]{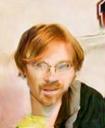}
\includegraphics[scale=0.30]{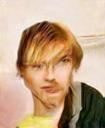}
\includegraphics[scale=0.30]{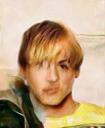}
\includegraphics[scale=0.30]{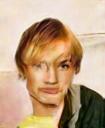}
\includegraphics[scale=0.30]{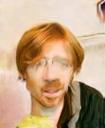}
\includegraphics[scale=0.30]{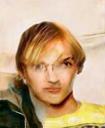}
\caption{One-shot face synthesis on two male face images. The leftmost two images are the input images and the rest of the images are generated images (128x128).}
\label{fig:one_shot_face_synthesis_p4}
\end{figure}
\begin{figure}
\centering
\includegraphics[scale=0.145]{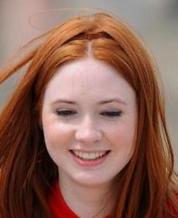}
\includegraphics[scale=0.145]{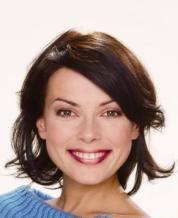}
\includegraphics[scale=0.145]{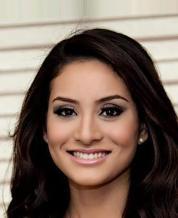}
\includegraphics[scale=0.2]{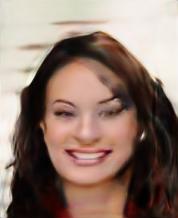}
\includegraphics[scale=0.2]{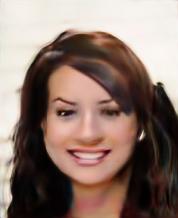} 
\includegraphics[scale=0.2]{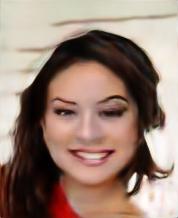}
\includegraphics[scale=0.2]{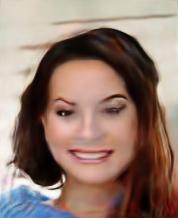}
\includegraphics[scale=0.2]{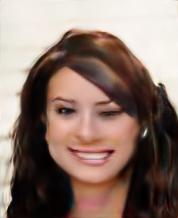}
\includegraphics[scale=0.2]{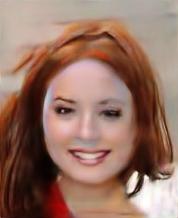}
\includegraphics[scale=0.2]{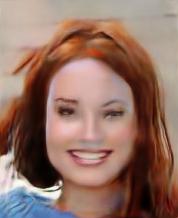}
\includegraphics[scale=0.2]{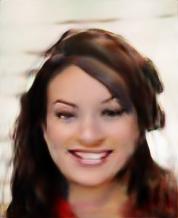}
\includegraphics[scale=0.2]{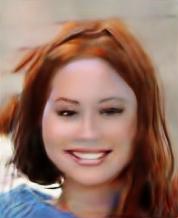}
\includegraphics[scale=0.2]{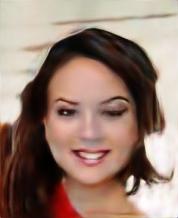}
\includegraphics[scale=0.2]{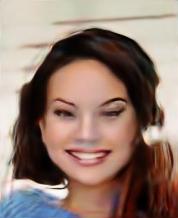}
\caption{Few-shot face synthesis using three face images. The top row leftmost three images are the inputs and the rest are generated images (256x256).}
\label{fig:few_shot_face_synthesis}
\end{figure}

We conducted image synthesis experiments for one-shot multi-class image synthesis on cat and dog faces for two inputs (refer Fig.\ref{fig:Dog2cat_p3_v11}), human faces for two inputs of male (refer Fig.\ref{fig:one_shot_face_synthesis_p4}) and female faces (refer Fig.\ref{fig:one_shot_face_composition_p9}), three inputs of female faces (refer Fig.\ref{fig:few_shot_face_synthesis}) and five inputs of flower images (refer Fig.\ref{fig:flower-101}). The generated images for the few-shot image synthesis for a selected set of ten images from the 100-shot-Obama dataset are in Figure \ref{fig:obama-10}. The LPIPS value for the same are given in Table \ref{tab:LPIPS_all_images} and Table \ref{tab:LPIPS_ip_gen}, where the top most row denotes the iteration number at which the images were generated using Mixdl, while for our method the number of stages are mentioned alongside training iterations that vary between 2-6k per stage.
We also report the FID values of our method compared to the baseline \cite{mixdl_2021} (Mixdl) in Table \ref{tab:FID_compared} and SIFID values of our method in Table \ref{tab:SIFID_values}. The SIFID values of mixdl are very large in comparison to our method and has been skipped in the table. 
The results of the generation of cat and dog images are given in Figure \ref{fig:Dog2cat_p3_v11}. The results of one-shot image synthesis on non-facial texture images of polka dot are given in Figure \ref{fig:few_shot_synthesis_polka_dotted_256}. 
\begin{table}[]
\centering
\caption{LPIPS metric computed for all generated images between consecutive image pairs for five inputs of flowers images, two inputs of male faces and female faces for the few-shot image synthesis task compared for Mixdl* (columns 2-5) and our method (columns 6-11).}
\resizebox{0.95\columnwidth}{!}{
\begin{tabular}{|c|c|c|c|c|c|c|c|c|c|c|} \hline
\textbf{LPIPS} & \textbf{30K*} & \textbf{40K*}  & \textbf{50K*} & \textbf{60K*} & \textbf{3000(6)} & \textbf{3500(6)} & \textbf{4000(6)} & \textbf{4500(6)} & \textbf{5000(6)} & \textbf{5500(6)}\\ \hline
Flowers-5 & -- & 0.48 & \textbf{0.39} & 0.53 & 0.59 & 0.59 & 0.58 & 0.59 & -- & --\\ \hline
Male faces & 0.41 & 0.38 & 0.36  & -- &  -- & -- & 0.28 & 0.28 & \textbf{0.26} & 0.27\\ \hline
Female faces & \textbf{0.27} & 0.32 & 0.32 & -- & -- & -- & \textbf{0.27}& \textbf{0.27} & \textbf{0.27} & \textbf{0.27}\\ \hline
\end{tabular}} 
\label{tab:LPIPS_all_images}
\end{table}

\begin{table}[]
\centering
\caption{LPIPS values computed between each input image and generated images for five inputs compared between Mixdl* (columns 2-4) and our method (columns 5-8). }
\resizebox{0.95\columnwidth}{!}{
\begin{tabular}{|c|c|c|c|c|c|c|c|c|} \hline
\textbf{LPIPS} & \textbf{30K*}  & \textbf{40K*}  & \textbf{50K*} & \textbf{60K*} & \textbf{3000(6)} & \textbf{3500(6)} & \textbf{4000(6)} & \textbf{4500(6)}\\ \hline
Flower-1 & -- &0.64 & \textbf{0.62} & 0.64 & 0.64 & 0.65 & 0.64 & 0.65\\ \hline
Flower-2 & -- & \textbf{0.52} & 0.44 & 0.63 & 0.63 & 0.62 & 0.62 & 0.63\\ \hline
Flower-3 & -- &0.66 & 0.65 & 0.67 & 0.64 & 0.64 & \textbf{0.63} & 0.64\\ \hline
Flower-4 & -- &0.63 & \textbf{0.60} & 0.64 & 0.69 & 0.68 & 0.69 & 0.69\\ \hline
Flower-5 & -- &0.68 & 0.66 & \textbf{0.51} & 0.63 & 0.63 & 0.63 & 0.63\\ \hline
Male-1 & 0.42 & 0.44 & 0.41 & --& 0.30 & \textbf{0.27} & 0.28 & 0.28\\ \hline
Male-2 & 0.47 & 0.43 & 0.44 & --& 0.30 & 0.29 & 0.31 & \textbf{0.28}\\ \hline
Female-1 & 0.33 & 0.36 & 0.37 & --& \textbf{0.26} & \textbf{0.26} & 0.27 & 0.27\\ \hline
Female-2 & 0.46 & 0.45 & 0.42 & --& 0.29 & \textbf{0.28} & \textbf{0.28} & \textbf{0.28}\\ \hline
\end{tabular}}
\label{tab:LPIPS_ip_gen}
\end{table}

\begin{table}[]
\centering
\caption{FID values between input images and the generated images for few-shot image synthesis task using Mixdl* (columns 2-4) and our method (columns 5-10).}
\resizebox{0.95\columnwidth}{!}{
\begin{tabular}{|c|c|c|c|c|c|c|c|c|c|} \hline
\textbf{FID} & \textbf{30K*}  & \textbf{40K*} & \textbf{50K*} & \textbf{3000} & \textbf{3500} & \textbf{4000} & \textbf{4500} & \textbf{5000} & \textbf{5500}\\ \hline
Flowers & -- & \textbf{200.94} & 219.63 & 244.73 & 236.52 & 238.56 & 244.33 & -- & --\\ \hline
Male faces & 238.03 & 205.94 & 201.05 & -- & -- & 217.78 & 202.74 & 205.14 & \textbf{189.16} \\ \hline
Female faces & 167.90 & 140.29 & 130.09 & -- & -- & 128.76 & 121.78 & \textbf{119.97} & 128.11 \\ \hline
\end{tabular}}
\label{tab:FID_compared}
\end{table}

\section{RESULTS AND ANALYSIS}
\label{Sec:RESULTS}
Initially, we considered ConSinGAN as a baseline for modelling multiple images, with a single input randomly selected in each iteration from the set of few images, keeping the image remained fixed throughout the critic operations, but the generated images were affected by mode collapse. The mode collapsed images on modelling two face images for three image pairs can be seen in Figure \ref{fig:mode_collapse}. We avoid the evaluation of mode collapsed images as all generated images are the same. We conjecture that mode collapse could be due to smaller batch sizes that doesn't consider all input images and the image remains fixed throughout the critic operations. 

\begin{table}[]
\centering
\caption{SIFID values between each input image and generated images for two inputs (rows 2-3, 4-5) and three inputs (rows 6-8) for few-shot image synthesis task using our method.}
\resizebox{0.9\columnwidth}{!}{
\begin{tabular}{|c|c|c|c|c|c|c|c|} \hline
\textbf{SIFID} & \textbf{3000}  & \textbf{3500} & \textbf{4000} & \textbf{4500} & \textbf{5000} & \textbf{5500}\\ \hline
Male-1 & -- & -- & 0.196 & \textbf{0.142} & 0.177 & 0.173 \\ \hline
Male-2 & -- & -- & 0.174 & 0.197 & 0.199 & \textbf{0.169} \\ \hline
2-Female-1 & -- & -- & 0.149 & \textbf{0.143} & 0.148 & 0.156 \\ \hline
2-Female-2 & -- & -- & 0.226 & \textbf{0.205} & 0.199 & 0.194 \\ \hline
3-Female-1 & \textbf{0.593} & 0.611 & 0.619 & 0.644 & -- & -- \\ \hline
3-Female-2 & 0.569 & \textbf{0.419} & 0.456 & 0.445 & -- & -- \\ \hline
3-Female-3 & \textbf{0.810} & 0.850 & 0.826 & 0.826 & -- & -- \\ \hline
\end{tabular}}
\label{tab:SIFID_values}
\end{table}

\subsection{Quantitative Results}
Table \ref{tab:FID_compared} compares the FID scores computed between the input images and the generated images. We can observe that our method has lower FID scores that implies better quality than the baseline for two inputs of human faces for both male and female faces. 
Table \ref{tab:LPIPS_ip_gen} compares LPIPS scores computed between each input image and the set of generated images for our method and Mixdl. We can observe that for 2-input case, our method scores better LPIPS scores than the baseline. For five inputs of flower images, our method falls behind the baseline for both LPIPS and FID scores. We conjecture that this could be due to fewer common correspondences and rough alignments of input images. It is easier to align correspondences with fewer and similar images but difficult when the number of classes are large and different, leading to less coherent samples. 
To summarize, training Mixdl is inefficient in a few-shot setting due to large number of network parameters and the checkpoint size. Table \ref{tab:SIFID_values} compares the SIFID values of our method for various input images. Values below 1 for SIFID scores indicate that the generated images are similar and share features from the input images. Since we have computed the SIFID scores of each input against all generated images, small SIFID scores imply that generated samples have features from multiple input images, which imply the transfer of visual attributes. The baseline method had very high SIFID scores, which could be due to poor attributes/features from multiple inputs.
\\

\subsection{Observations and Analysis}
The selection of input images becomes crucial for data efficient few-shot GANs, which are important for the decision boundary of the discriminator. The abrupt changes in the output of the generator is due to the discontinuities in latent space and possible reason for degradation of few-shot GANs. 
The assumption for novel image synthesis is that the generated image should have the similar global layouts as that of original images with possible attribute transfer from other input images.
\par
From the results of few-shot face synthesis, we can observe that our method is able to faithfully generate diverse set of generated images. Our method also extends to non-facial classes like texture images or flower class. Also, our method can extend to related classes as one-shot multi-class image synthesis, as in the case of dog and cat images. We can observe that the generated images with largest diversity were the ones with similarity in textures, shapes and color, which is observed in the case of polka-dot texture images. 
We can infer that neither FID and SIFID are good evaluation metrics for one-shot multi-class image synthesis. FID computation requires lot more input images and gives large FID values when the  number of input images are a few, while the generated images are large in number. On the other hand, SIFID is suitable for a single image and doesn't take into consideration for multiple input images and feature transfer. 
As a limitation in comparison to other GAN methods, the generated samples of our method also tend to have non-smooth interpolations to other samples. One can always find some set of input images that are inherently difficult for our method to generate leading to reduced semantics.  

\section{\uppercase{Conclusion and Future Scope}}
\label{sec:conclusion}
In this work, we improved the capabilities of single image models to accommodate multiple images. This is possible with simple assumptions of similarities in underlying content and a modified discriminator architecture and objective function. When we consider two face images that are roughly aligned, but differ in other aspects like texture, color and light intensities, our method involves learning a distribution of the patches that appear from the natural composition of the input images. The idea extends to multiple images, assuming that the images are roughly aligned and the images share similar underlying content layouts. Our method generates diverse set of hundreds of data samples by training on just two input images. Future work can focus on improving control over the style at global and local level.

\bibliographystyle{apalike}

{\small
\bibliography{example}}
\end{document}